\documentclass{article}

\usepackage{PRIMEarxiv}

\usepackage[utf8]{inputenc} 
\usepackage[T1]{fontenc}    
\usepackage{hyperref}       
\usepackage{url}            
\usepackage{booktabs}       
\usepackage{amsfonts}       
\usepackage{nicefrac}       
\usepackage{microtype}      
\usepackage{lipsum}
\usepackage{fancyhdr}       
\usepackage{graphicx}       
\usepackage{color}
\usepackage[normalem]{ulem} 
\graphicspath{{media/}}     

\title{ConTEXTual Net: A Multimodal Vision-Language Model for Segmentation of Pneumothorax}

\author{
 Zachary Huemann \\
  University of Wisconsin-Madison\\
  Madison, WI 53705 \\
  \texttt{zhuemann@wisc.edu} \\
   \And
 Xin Tie \\
  University of Wisconsin-Madison\\
  Madison, WI 53705 \\
  \texttt{xtie@wisc.edu} \\
  \And
   \And
 Junjie Hu \\
  University of Wisconsin-Madison\\
  Madison, WI 53705 \\
  \texttt{junjie.hu@wisc.edu} \\
  \And
 Tyler J. Bradshaw \\
   University of Wisconsin-Madison\\
   Madison, WI 53705 USA\\
  \texttt{tbradshaw@wisc.edu} \\
}

\begin{document}
\maketitle

\begin{abstract}
Radiology narrative reports often describe characteristics of a patient's disease, including its location, size, and shape. Motivated by the recent success of multimodal learning, we hypothesized that this descriptive text could guide medical image analysis algorithms. We proposed a novel vision-language model, ConTEXTual Net, for the task of pneumothorax segmentation on chest radiographs. ConTEXTual Net utilizes language features extracted from corresponding free-form radiology reports using a pre-trained language model. Cross-attention modules are designed to combine the intermediate output of each vision encoder layer and the text embeddings generated by the language model. ConTEXTual Net was trained on the CANDID-PTX dataset consisting of 3,196 positive cases of pneumothorax with segmentation annotations from 6 different physicians as well as clinical radiology reports. Using cross-validation, ConTEXTual Net achieved a Dice score of 0.716$\pm$0.016, which was similar to the degree of inter-reader variability (0.712$\pm$0.044) computed on a subset of the data. It outperformed both vision-only models (ResNet50 U-Net: 0.677$\pm$0.015 and GLoRIA: 0.686$\pm$0.014) and a competing vision-language model (LAVT: 0.706$\pm$0.009). Ablation studies confirmed that it was the text information that led to the performance gains. Additionally, we show that certain augmentation methods degraded ConTEXTual Net's segmentation performance by breaking the image-text concordance. We also evaluated the effects of using different language models and activation functions in the cross-attention module, highlighting the efficacy of our chosen architectural design. 
\end{abstract}


\section{Introduction}
\label{sec:introduction}
Radiology is a multimodal field. Picture archiving and communication systems (PACS) contain medical images as well as accompanying reports generated by radiologists. These reports serve as the official record of the reading physicians' interpretations for radiological exams, playing an essential role in communicating findings to patients and their healthcare teams. Additionally, such reports offer radiologists invaluable context regarding prior imaging results when they are interpreting follow-up exams. For example, in reading a current set of images, radiologists often review the patient's prior images and reports to ascertain the location and extent of the disease. This allows for monitoring disease evolution over time and assessing the effectiveness of treatments. The comparative review also helps in identifying new developments or subtle changes that might be overlooked in the absence of a reference point. Although reviewing past exams can be time-consuming, its value in many diagnostic applications is undeniable. 

A clinical scenario where repeated images are acquired is the management of patients diagnosed with pneumothorax. Pneumothorax is a condition in which air accumulates in the space between the lung and chest wall. Due to its life-threatening nature, rapid detection and intervention are crucial to prevent severe morbidity or mortality. If the pneumothorax is large or increasing in size,  a chest tube must be inserted \cite{zarogoulidis_pneumothorax_2014}. Therefore, monitoring its changes is important. Pneumothorax can be difficult to detect and may not be obvious on follow-up images. As a result, physicians often consult prior images and reports as part of their standard workflow. This workflow could be greatly enhanced with the aid of automatic pneumothorax segmentation tools. 

Motivated by the success of recent multimodal vision-language models  \cite{ramesh_hierarchical_2022} \cite{saharia_photorealistic_2022}, we aim to leverage information in clinical reports to improve medical image analysis. Models like ConVIRT \cite{zhang_contrastive_2022} and GLoRIA \cite{huang_gloria_2021} have used image-report pairs and contrastive learning objectives to learn vision representations, which have shown promising results in downstream classification tasks. However, these models only utilized text for pre-training, without integrating it into the model to guide image analysis.  Other approaches employed image and text encoders to identify instances of pneumonia and placed bounding boxes around them \cite{Bhalodia_multimodal_bounding_box_2021}, but did not provide pixel-level segmentation. LAVT \cite{yang_lavt_2022} produced pixel-wise predictions from image text pairs but was developed for referring image segmentation of household items rather than medical images. LVIT \cite{li_lvit_2022} is a vision-language model that took chest x-ray (CXR) images and text as inputs and generated segmentation masks for patients diagnosed with coronavirus disease 2019 (COVID-19). Nevertheless, the text inputs were synthesized from the ground-truth labels instead of using real free-form radiology reports.

In this work, we aimed to apply multimodal learning to integrate real physician reports into the task of pneumothorax segmentation on chest radiographs. To this end, we developed an algorithm that maps concepts from language space into medical image space so that descriptive text can be used to guide image segmentation. Instead of bounding box detection, we pursued fine-grained segmentation, as it provides delineation of boundaries, and allows for morphological analysis and quantitative measurements (e.g., volume) of the disease. Compared to non-medical segmentation tasks, pneumothorax segmentation poses distinct challenges, including limited image availability, and the expertise and overall cost required for annotation. Our approach of directly incorporating physician-generated text into the image analysis has the benefit of improving segmentation accuracy by allowing the model to leverage the physician's expertise. Moreover, this integration of language paves the way for real-time, physician-guided disease assessment. 

The rest of the paper is organized as follows: section two covers the methodology behind our proposed vision-language model, named ConTEXTual Net. Section three details the experimental setup. We present the segmentation results of our model and ablation analysis in section four. Lastly, we discuss our findings and conclusions. 

\section{Methods}

\subsection{Model Architecture}

\begin{figure*}[!h]
\centerline{\includegraphics[width=0.95\textwidth]{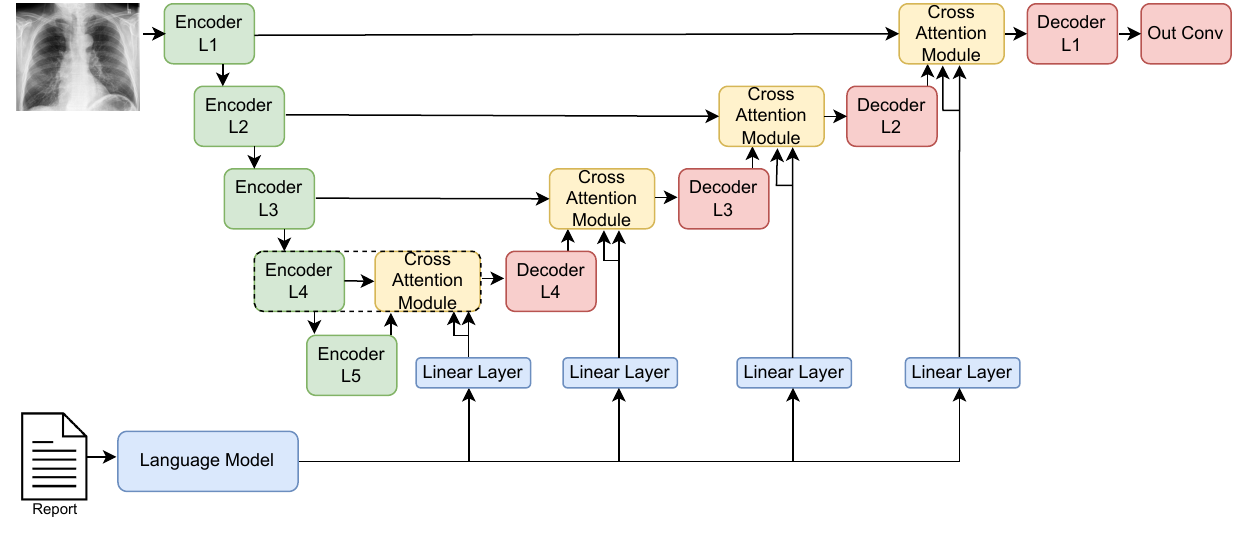}}
\caption{ConTEXTual Net combines a U-Net and a transformer architecture. It uses the encoder layers of the U-Net to extract visual representations and uses a pretrained language model (\textbf{bottom-left}) to extract language representations. It then performs cross-attention between the modalities, and finally uses the decoder layers (\textbf{top-right}) of the U-Net to predict the segmentation masks. The cross-attention module (dotted box) is further detailed in Fig~\ref{fig2}.}
\label{fig1}
\end{figure*}

ConTEXTual Net's architecture can be seen in Fig.~\ref{fig1}, and the implementation details can be found in the open source project\footnote{\url{https://github.com/zhuemann/ConTEXTualSegmentation}}. Vision features are extracted from the image via a U-Net \cite{long_fully_2015} encoding scheme, shown in green, and language features are extracted via a pretrained language model, shown in blue. This approach leverages the ability of the U-Net to contour disease and the ability of transformer-based language models to extract semantically rich vectors which can be used to localize the disease. Within the U-Net, each vision encoder layer is a stack of two sub-layers, where each sub-layer is a convolution followed by batch normalization and ReLU activation. The output of each encoder layer is fed as a skip connection to a cross-attention module (seen in yellow in Fig.~\ref{fig1} and described in Section~\ref{sec:cross-attetnion}). Meanwhile, the encoder output is also downsampled via max pooling and fed to the next encoder layer. The output of the cross-attention module is subsequently fed to a decoder layer which is also a stack of two convolution sub-layers with batch normalization and ReLU (similar to the encoder layer) and then upsampled as inputs to the next cross-attention module. The last layer of the decoder is a $1\times1$ convolution, which reduces the channels to a single output channel, and is used for pixel-level prediction.

To integrate language into the model, a pre-trained language model is used to encode a text report into contextualized text embeddings. These embeddings are further projected by a linear layer and fed as inputs to the cross-attention modules.
 The model is trained using supervised learning with a cross-entropy loss on the prediction and ground-truth segmentation labels. The pre-trained language model is frozen during training, which reduces the number of parameters that need to be learned. 

\subsection{Language Cross-Attention}
\label{sec:cross-attetnion}
ConTEXTual Net uses a cross-attention module to inject the embeddings from the language model into the vision-based segmentation model. The text embeddings are injected into the decoding side of a U-Net. Conceptually, the text embeddings contain semantic information about the presence and location of the disease and can be used to guide the U-Net segmentation model. The cross-attention between the text embeddings and the decoded feature maps produces a pixel-wise attention map. This pixel-wise attention map then gets fed into a \emph{Tanh} activation function to normalize values between $-1$ and $1$. The normalized pixel-wise attention map is then multiplied pixel-wise with the query feature map. The pixel-wise attention map, $\mathbf{A}$, is obtained by

\begin{equation}
    \mathbf{A} =\emph{softmax}\left(\frac{(\mathbf{\bar{Q}}\mathbf{W}^Q(\mathbf{K}\mathbf{W}^K)^\top)}{\sqrt{d_k}}\right)\mathbf{V}\mathbf{W}^V \label{eq1}
\end{equation}
where $\mathbf{A}$ is the pixel-wise attention map, $\mathbf{\bar{Q}} \in \mathbb{R}^{w h \times c}$ is the query vectors flattened from the upsampled feature map $\mathbf{Q}\in \mathbb{R}^{w \times h \times c}$ of size $w\times h$ in $c$ channels, $\mathbf{K}=\mathbf{V}\in\mathbb{R}^{l \times d_p}$ are the projected text embeddings of a report of $l$ words, and $\mathbf{W}^Q$, $\mathbf{W}^K$, and $\mathbf{W}^V$ are the learnable weights that project the query, key and value vectors to the same dimensional space. Note here the projected text dimension $d_p$ is chosen to match the number of channels $c$ at each decoder layer of the U-Net. Subsequently, the cross-attention output is calculated as 


\begin{equation}
    \mathbf{Q}^* = \emph{tanh}(\mathbf{A})*\mathbf{Q}\end{equation}
where $*$ is the element-wise multiplication of the input feature map and the normalized pixel attention map, and $Q^*$ is the attention-weighted feature map. Empirically, \emph{Tanh} activation was found to perform better than other activation functions.  Single-headed attention is used due to computational constraints of attending between every pixel and each token output from the language model.

\begin{figure*}[!h]
\centerline{\includegraphics[width=0.9\textwidth]{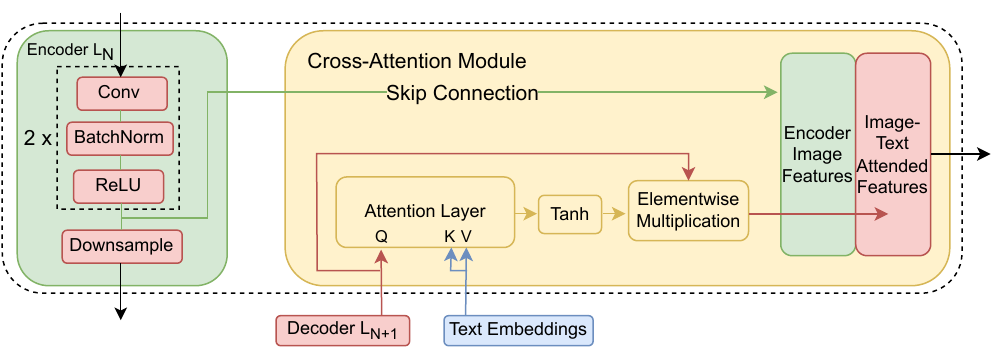}}
\caption{The cross-attention module takes in the upsampled feature maps (\textbf{bottom}) which are used as the query, and the projected word-level embeddings from the language model, which are used as both the key and value. It calculates pixel-level attention maps which are used to weight the decode feature maps.}
\label{fig2}
\end{figure*}

\subsection{Augmentations}
Data augmentation is the process of altering training data to synthetically increase dataset size so that the model better generalizes to new situations. For multimodal models, these data augmentations must preserve the concordance between the text description and image. For example, if horizontal image flips are used, the descriptions of ``left" and ``right" no longer correspond to the image. This can be addressed by either picking augmentations that are invariant to the other modality or augmenting both modalities to retain concordance. In this work, we focused on finding augmentations that are invariant to the other modality.

We considered the following set of image augmentations: horizontal flip 50\% of the time, 30\% of the time choosing one from RandomContrast, RandomGamma, and RandomBrightness, and again 30\% of the time choosing from ElasticTransform, GridDistortion, and OpticalDistortion and lastly ShiftScaleRotate, all of which have been used previously for pneumothorax segmentation \cite{Aimoldin2019} and \cite{abedalla_chest_2021}. Out of those, horizontal flipping was the only augmentation that we found to break the image-text concordance and was thus left out from all experiments unless otherwise stated. All augmentations were implemented using the \texttt{Albumentations} library \cite{buslaev_albumentations_2020}.


Text augmentations were used to improve the model's generalizability to different writing styles. Specifically, two augmentations were used: sentence shuffling and synonym replacement. In sentence shuffling, the text is split into sentences and randomly rearranged. For radiology reports, sentences are generally self-contained with few inter-sentence dependencies, and we expect sentence shuffling to have little to no effect on the meaning. For synonym replacement, we used RadLex \cite{langlotz_radlex_2006}, a radiology ontology that contains definitions and synonyms for radiology-specific terms. During train time, each word in the report that had a RadLex-listed synonym was replaced with the synonym 15\% of the time. 

\section{Experimental Setting}

\subsection{CANDID-PTX Dataset}

We developed and evaluated ConTEXTual Net using the CANDID-PTX dataset. The CANDID-PTX dataset consists of 19,237 chest radiographs with reports and segmentations of pneumothoraces, acute rib fractures, and intercostal chest tubes.  We focused on pneumothorax in this study. There are 3,196 positive cases of pneumothorax with segmentation annotations labeled by 6 different physicians. It is important to note that the physician who dictated the original report was different from the physician responsible for labelling. 

The segmentation performance is evaluated using Dice scores given by 
\begin{equation} Dice = \frac{2|A \cap B|}{|A| + |B|} \end{equation}
which compares two observers' predictions (i.e., model vs annotation or one physician annotator vs another). The inter-rater Dice similarity scores between the 6 physicians ranged from 0.64-0.85. The primary annotator labeled 92.7\% of the images in the dataset and had a mean Dice similarity score of 0.712 when compared to the other 5 physicians. The dataset was originally collected with approval from an ethics committee with waiver of informed consent and was made available under a data use agreement \cite{feng_curation_2021}.  

\subsection{Model Comparison}
To show the efficacy of our architectural design, we compared ConTEXTual Net against methods from previous studies. Specifically, we trained a U-Net model \cite{ronneberger_u-net_2015} that shares the same architecture with the vision-only component of ConTEXTual Net. We also included a baseline model that was build on the standard U-Net, with the encoder being replaced by Resnet50\cite{he_deep_2015}. Additionally, we compared against a U-Net with the encoder weights initialized from the GLoRIA model, which was pre-trained using multimodal contrastive learning with approximately 200k image-report pairs of chest x-rays \cite{huang_gloria_2021}. Lastly, we finetuned LAVT \cite{yang_lavt_2022}, which is a state-of-the-art vision-language model, for the task of text-guided pneumothorax segmentation. LAVT uses a Swin transformer to encoder the image information, a BERT model to encode the language, and fuses these via a pixel-word attention module.

\subsection{Language Models}

ConTEXTual Net uses language models to extract semantically important statements from the reports. In this study, we compared T5-Large's encoder \cite{raffel_exploring_2020}, RoBERTa-Large \cite{liu_roberta_2019}, RadBERT \cite{yan_radbert_2022} and BERT \cite{devlin_bert}, as the language encoders. T5-Large is a larger model which contains 770M parameters, RoBERTa-Large contains 354M parameters, and RadBERT and Bert are considerably smaller at 125M and 110M parameters respectively. All models are transformer-based but their training datasets and tasks differ. T5 is trained on the c4 dataset which contains roughly 300G of text and is intended to be capable of diverse tasks, including text classification, question answering, machine translation, and abstractive summarization. RoBERTa-Large and Bert are trained via masked language modeling but RoBERta-Large was trained on a 161G text corpus whereas BERT's training corpus was 16GB. RadBERT is initialized from RoBERTa-Base but is trained with roughly 4M additional radiology reports from the U.S. Department of Veterans Affairs. 

The reports were fed into the language models and the hidden state vectors were used as our report representations. The report representation has dimensions of $512\times1024$ (token length $\times$ embedding dimension) for T5-Large and RoBERTa-Large and $512\times768$ for RadBERT and BERT. This report representation then goes through a projection head to lower the embedding dimensionality of the hidden state vector to match the number of channels in the encoder feature maps. This is a necessary step for the language cross-attention to work at multiple levels in the U-Net decoder. Language models were imported via \texttt{HuggingFace} library \cite{wolf_transformers_2020}. 

\subsection{Ablation Analysis}

We performed ablations studies to determine the additive value of augmentations as well as ConTEXTual Net's components. Ablation studies were performed in three settings: without any augmentations, with image augmentations only, and with both image and text augmentations. We additionally tested ConTEXTual net with the full language encoder and cross-attention modules but the input text was replaced with an empty string to help quantify the effects of the physician text. RoBERTa-Large, RadBERT, and Bert were swapped with the T5 encoder to examine how sensitive ConTEXTual Net is to the language model used. To determine the best activation function in the cross-attention module we tested using no activation function, ReLU, Sigmoid, and the hyperbolic tangent function. Along with these experiments we examined the impact of only using a single attention module at the four different levels of the network. This is meant to probe the question of whether textual information should be inserted early in the image analysis or at later stages. All language model, activation function, and attention module integration experiments were performed with vision augmentations only. 

\subsection{Model Hyperparameters}
All ConTEXTual Net models were trained with the AdamW optimizer, a learning rate of 5e-5, and 100 epochs with a binary cross-entropy loss. The native image dimensions of $1024\times 1024$ were used. The model which did best on the validation set was used on the cross-validation test set. We report the average and standard deviation of 5-fold Monte Carlo cross-validation. All models were trained on NVIDIA A100 GPUs.

\section{Results}
\subsection{Model Comparison}

\begin{table} [t]
\centering
\caption{Model Comparisons}
\label{fig:results}
\begin{tabular}{cll}
\toprule
Model Type  & AVG Dice   & SD \\
\hline
\midrule
ConTEXTual vision only U-Net & 0.680 & 0.014\\
Resnet50 U-Net & 0.677 & 0.015\\
GLoRIA         & 0.686 & 0.014\\
LAVT           & 0.706 & 0.009\\
ConTEXTual Net & \textbf{0.716}  & 0.016\\
Primary Physician Annotator & 0.712 & 0.044\\
\bottomrule
\end{tabular}
\end{table}

\begin{figure*}[!t]
\centerline{\includegraphics[width=480pt]{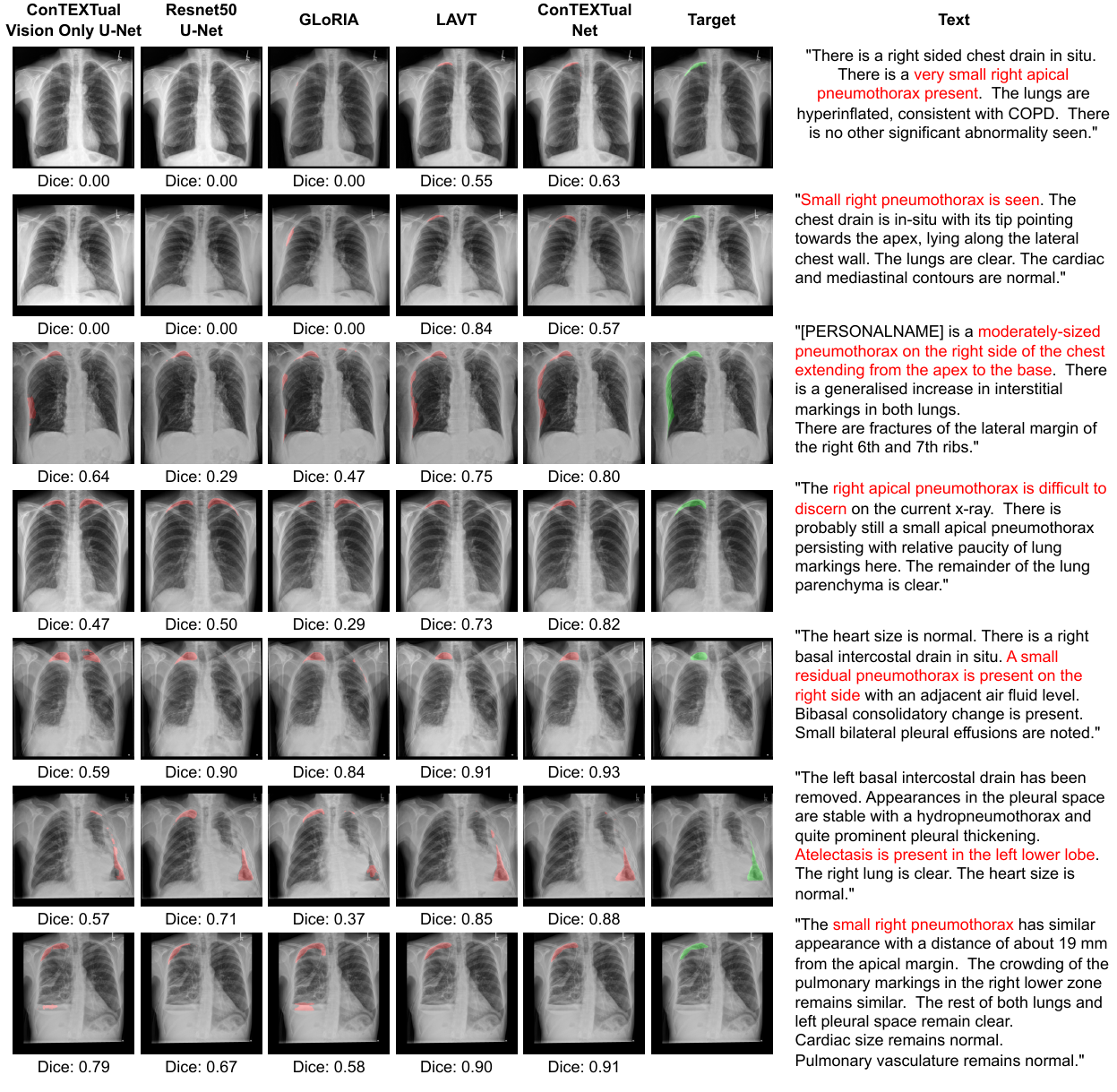}}
\caption{Predictions from all five evaluated models as well as the physician labeled target and a portion of the text is shown. Note how the model used the text to detect subtle cases (top two cases), and can use descriptions like "extending from the apex to the base" (third case) to better segment disease. Text also can help it avoid false positives (cases 4-7).}
\label{model_comparison}
\end{figure*}

Table \ref{fig:results} shows a comparison of Dice scores for all models and the primary physician annotator as compared to the other physician annotators. Overall, the best-performing configuration of ConTEXTual Net (Dice 0.716 $\pm$ 0.016) outperformed the baseline U-Net (0.680 $\pm$ 0.014) and LAVT (0.706 $\pm$ 0.009) and performed similarly to the primary labeling physician (0.712 $\pm$ 0.044). Example images with results are shown in Fig.~\ref{model_comparison}.

We report results from ablation analyses in Table \ref{fig:ablation}. Due to the joint dependencies of ConTEXTual Net's components and the data augmentations used during training, we report results separately based on the types of data augmentations.

\subsection{Vision Augmentations}

We evaluated the added value of ConTEXTual Net's cross-attention module and different vision-based augmentation methods. Without vision augmentations, ConTEXTual Net (0.668 $\pm$ 0.010) only slightly outperformed the baseline U-Net (0.649 $\pm$ 0.014). When the vision augmentations were applied, ConTEXTual Net's (0.716 $\pm$ 0.016) relative performance compared to the baseline U-Net (0.680 $\pm$ 0.014) increases to a 3.6 Dice point improvement. The performance of ConTEXTual Net significantly decreases (0.671 $\pm$ 0.019) when the reports are replaced with padding tokens (i.e., an empty string was used as input), which indicates that the reports are indeed helping guide the segmentation. When horizontal image flipping is applied, ConTEXTual Net's (0.675 $\pm$ 0.016) performance decreases by 4.1 Dice points and is comparable to using the empty string as input. This suggests the model learned to ignore the text when augmentations break the image-text correspondence. All other vision augmentations improved the model performance.

\subsection{Text Augmentations}

Although text augmentation represents a promising approach to vision-language data augmentations, text augmentations did not lead to gains in segmentation performance as shown in Table \ref{fig:ablation}. Using RadLex-based synonym replacement (0.705 $\pm$ 0.008) resulted in a decrease in performance of 1.1 dice points. Sentence shuffle (0.713 $\pm$ 0.023) led to increased variance across runs. Using both augmentations with the T5 encoder (0.714 $\pm$ 0.014) had very little effect on the model's performance. 

\begin{table}
\centering
\caption{Ablation Study of ConTEXTual Net}
\label{fig:ablation}
\resizebox{\columnwidth}{!}{%
\begin{tabular}{cll}
\toprule
Model Type  & AVG Dice   & SD \\
\hline
\midrule
No Augmentations & &\\
\cmidrule[1pt]{1-1}
Baseline U-Net & 0.649 & 0.014 \\
ConTEXTual Net & 0.668 & 0.010 \\
\midrule
Vision Augmentations & & \\ 
\cmidrule[1pt]{1-1}
Baseline U-Net & 0.680 & 0.014\\
ConTEXTual Net & \textbf{0.716} & 0.016\\
ConTEXTual Net with flipping & 0.675 & 0.016\\
ConTEXTual Net w/o reports & 0.671 & 0.019\\
\midrule
Text Augmentations & & \\ 
\cmidrule[1pt]{1-1}
w/o text augmentations & \textbf{0.716} & 0.016 \\
Synonym Replacement & 0.705 & 0.008 \\
Sentence Shuffle & 0.713 & 0.023 \\
Synonym + Sentence Shuffle & 0.714 & 0.014\\
\midrule
  & & \\
  & & \\
\end{tabular}
\quad
\begin{tabular}{cll}
\toprule
Model Type  & AVG Dice   & SD \\
\hline
\midrule
Language Models & & \\ 
\cmidrule[1pt]{1-1}
ConTEXTual Net (T5) & \textbf{0.716} & 0.016 \\
ConTEXTual Net (RoBERTa-Large) & 0.713 & 0.010\\
ConTEXTual Net (RadBERT) & 0.716 & 0.022\\
ConTEXTual Net (BERT) & 0.713 & 0.020\\

\midrule
Activation Functions & & \\
\cmidrule[1pt]{1-1}
ConTEXTual Net (Tanh) & \textbf{0.716} & 0.016 \\
ConTEXTual Net (ReLU) & 0.698 & 0.027 \\
ConTEXTual Net (Sigmoid) & 0.710 & 0.010 \\
ConTEXTual Net (No Activation) & 0.704 & 0.011 \\
\midrule
Cross-attention Integration & & \\
\cmidrule[1pt]{1-1}
Attention Module L4 & 0.712 & 0.019\\
Attention Module L3 & 0.709 & 0.013\\
Attention Module L2 & 0.685 & 0.021\\
Attention Module L1 & 0.679 & 0.011\\
\bottomrule
\end{tabular}
}
\end{table}

\subsection{Language Models}

We evaluated using RoBERTa-Large (0.713 $\pm$ 0.010), RadBERT(0.716 $\pm$ 0.022), and BERT (0.713 $\pm$ 0.020) instead of T5 as the language encoder while using vision augmentations. This had a negligible impact on performance, suggesting that ConTEXTual Net is robust to the language model used.


\subsection{Activation Functions}

The use of different activation functions was investigated using T5 as our language encoder with vision augmentations and without text augmentations. It was found that the hyperbolic tangent activation function (0.716 $\pm$ 0.016) performed the best outperforming the sigmoid function (0.710 $\pm$ 0.010). Using ReLU was found to decrease performance (0.698 $\pm$ 0.027) when compared to not using any activation function (0.704 $\pm$ 0.018). 

\subsection{Integration of Cross-attention Modules at Different Levels}

We found that the lower the attention was integrated the better the model performed. The L4 (i.e., decoder layer 4) cross-attention module (0.712 $\pm$ 0.019) performed slightly better than the L3 cross-attention module (0.709 $\pm$ 0.013). Moving from the L3 to L2 cross-attention module (0.685 $\pm$ 0.021) precipitated the largest drop in performance of 2.4 dice points. Finally, the L1 cross-attention module performed slightly worse than the L2 cross-attention module.

\begin{figure*}[!t]
\centerline{\includegraphics[width=430pt]{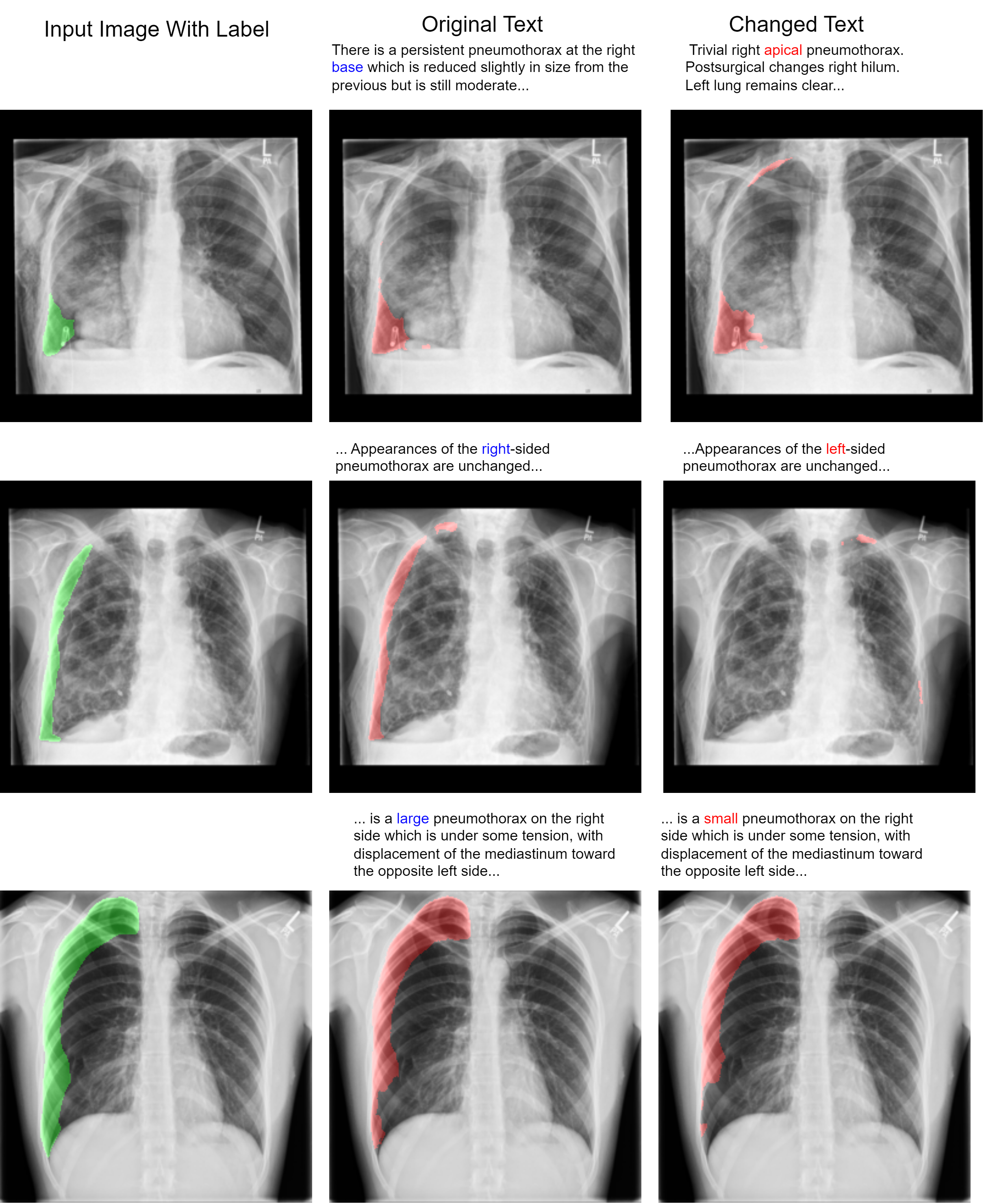}}
\caption{The same input image with different text is fed into the multimodal model. In the top row, an incorrect report describing an apical pneumothorax is used as input with an image, demonstrating that location descriptors like ``apical'' and ``base'' carry relevant information for segmentation. In the middle row, we show an example of an image and text with the term ``right'' changed to ``left''. This illustrates the model's sensitivity at the word level. In the bottom row, we changed the term ``large'' to ``small'', which resulted in a  reduction of segmented pixels by 10\%. Note ``left'' and ``right'' correspond to the patient's ``left'' and ``right''.  }
\label{changed_text}
\end{figure*}

\begin{figure*}[!t]
\centerline{\includegraphics[width=500pt]{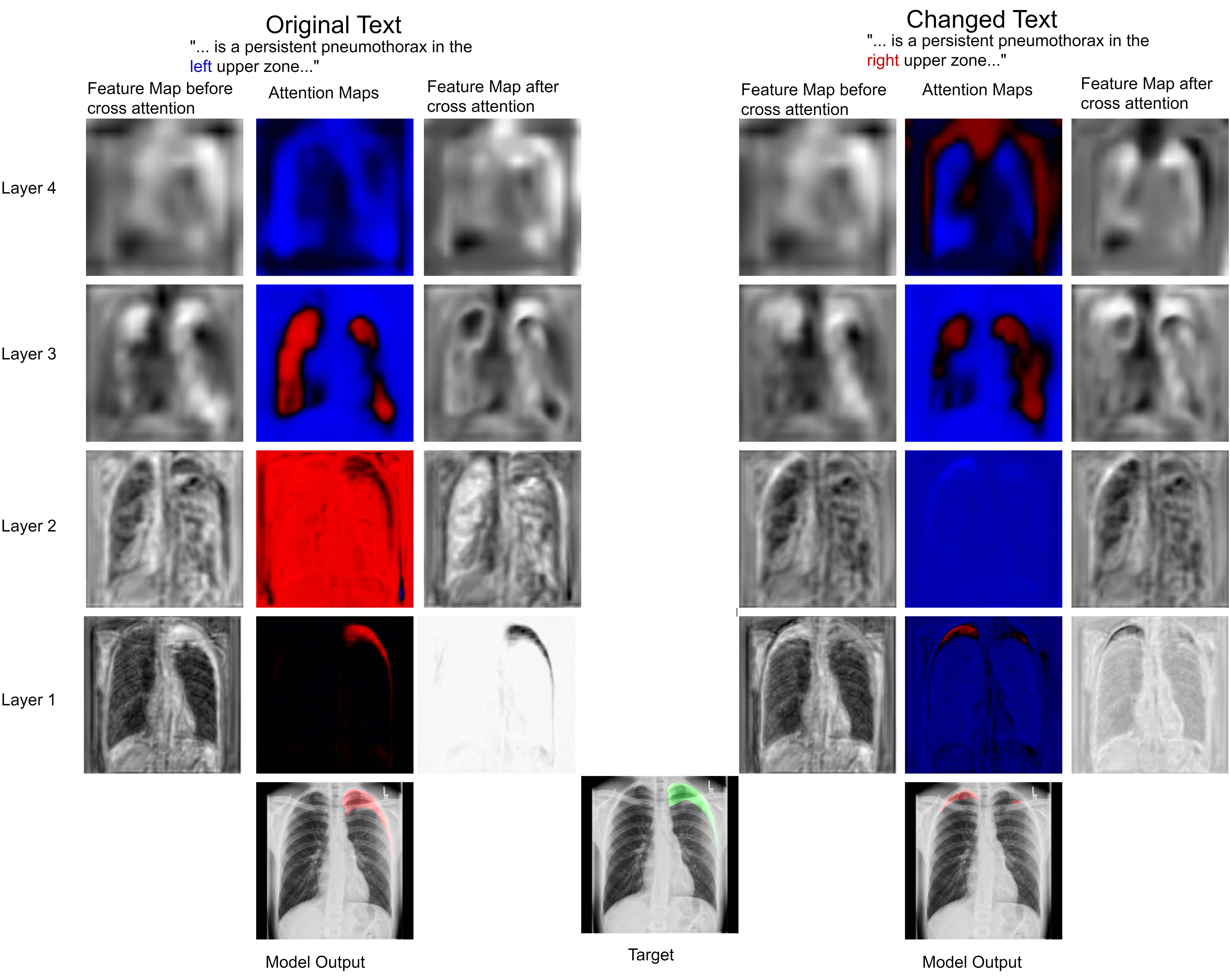}}
\caption{The same image is fed to the model with only a single word changed in the text report: ``left'' was switched to ``right''. Feature maps of the same channels at different layers of the U-Net are shown when two different text inputs are used. Shown are the input feature maps to the language cross-attention module, the attention output by the cross-attention module after the \emph{Tanh} activation, and the feature maps after the pixel-wise multiplication. In this case, it can be seen how the language changes the attention maps and guides the feature maps to reflect this change in the language. In the original text, the attention maps suppress the pixels in the right lung and portions of the left lower zone. In contrast, the attention maps in the case of the altered text suppress the pixels in the left lung but fail to completely suppress this signal and the result is a prediction of both a right and a small left pneumothorax.}
\label{attentionMaps}
\end{figure*}

\section{Discussion}

In this work, we proposed a method to extract information from clinical text reports and integrate it into a medical image segmentation algorithm. Our multimodal approach led to a 3.6 Dice point improvement over a traditional vision-only U-Net and achieved accuracy that matched physician performance. Additionally, we demonstrated the importance of maintaining image-text concordance when performing data augmentation and show that early fusion leads to more accurate predictions for multimodal segmentation.

 We showed the feasibility of using language from radiology reports to guide the output of segmentation algorithms. To illustrate how text can guide pixel-level predictions, we can change the input text and observe how the segmentation model reacts. For example, in Fig.~\ref{changed_text}, we changed the word ``right" to ``left", ``large pneumothorax" to ``small pneumothorax", and the location descriptor from ``base" to ``apical", and observed corresponding changes in the output segmentation map. We also showed that flipping an image during data augmentation can negatively impact model performance. In Fig.~\ref{attentionMaps}, we show that changing the text from ``left" pneumothorax to ``right" pneumothorax alters the attention maps from the cross-attention module, resulting in changes in the corresponding feature maps and model output.

ConTEXTual Net compared favorably to other vision-language models. The GLoRIA model, which only used text reports for pre-training, performed similarly to the baseline U-Net. This is not unexpected, as the purported advantage of GLoRIA is its performance in low-label settings, with limited advantages when datasets are sufficiently large \cite{huang_gloria_2021}. Our model slightly outperformed LAVT. LAVT was developed for segmenting the object(s) described in the text caption. Text captions were on average 1.6 words to 8.4 words long in the LAVT study, depending on the dataset. In contrast, our study had a single segmentation target, which was pneumothorax, and the physician-generated text that described the disease was on average much longer at 63.7 words. Another key difference in methods is our use of a CNN as the vision encoder, whereas LAVT used a Swin vision transformer. This design choice is driven by the lower number of images in medical databases compared to the natural image databases LAVT is trained on and the observation that CNNs are typically more sample efficient \cite{dosovitskiy_image_2021}. 

 
Our cross-attention integration experiments reveal a key insight into the integration of language within the vision encoder. Notably, we observed that integrating language at the lower levels of the U-Net improved model performance. This suggests that prioritizing early integration may be a strategic design choice. Cross-attention at a single layer may be a pragmatic approach to optimize resource utilization.

ConTEXTual Net was insensitive to the language model used to encode the radiology reports, but it was sensitive to the data augmentation methods and to the activation function used within the cross-attention module. We found that Tanh, by both bounding the output of the cross-attention and preserving the negative values, performed better than other activation functions in the cross-attention module. This may be due to the fact that the language model was frozen and not able to adapt to changes in activation function. LAVT likewise found that Tanh was the best activation function \cite{yang_lavt_2022}.

Multimodal medical image segmentation algorithms have several potential applications in radiology workflows. A primary motivation for this work was to address the challenge of reviewing follow-up imaging exams. For example, many patients who present with pneumothorax get an initial chest X-ray and then receive another chest X-ray within 3-6 hours to monitor the pneumothorax  \cite{li_pneumothorax_2014}. This means radiologists must visually compare the size and extent of the disease on previous chest X-rays. Multimodal models such as ConTEXTual Net could help with tasks involving longitudinal assessment. Vision-language segmentation models may also enable segmentation based on physician dictation, which could enable voice-guided disease quantification. While a practical limitation of ConTEXTual Net is that it requires language as input, which precludes its use on exams that have not yet been reviewed by a physician, it does provide a mechanism by which physicians can work together with AI to produce better outputs. These types of models can help to address patient and clinician concerns about the use of autonomous AI in medicine.  

In this study, we only analyzed positive cases of pneumothorax. Prior studies on pneumothorax segmentation included both positive and negative cases. For example, in the SIIM-ACR Pneumothorax Segmentation challenge, models that erroneously placed regions of interest in cases that were negative for pneumothorax were assigned a Dice score of 0. Models that correctly refused to place regions of interest in negative cases were assigned a Dice score of 1 \cite{abedalla_chest_2021}. Since the SIIM-ACR dataset had mostly negative cases, simply predicting negative on all cases resulted in a Dice score of around 0.79. We did not include negative cases because the input text to our multimodal model explicitly states whether the case is negative or positive. A language model could easily classify the cases as positive or negative based on the report  \cite{feng_curation_2021}. Therefore, it isn't appropriate to compare our multimodal model, which has access to the report text, to other vision models that don't have a way to use the text. Consequently, it is difficult to directly compare the results of our multimodal study to other vision-only studies on pneumothorax segmentation. Instead, we compared ConTEXTual Net's performance to the degree of inter-physician variability on the same dataset. We found that ConTEXTual Net's segmentation accuracy was comparable to the variability between physician contours.


There are several limitations of this proof-of-concept study. First, the study was trained and evaluated using a single dataset. This limits our ability to know how generalizable the ConTEXTual Net architecture is to other related tasks. Despite the availability of large medical imaging datasets that have images and radiology reports (e.g., MIMIC-CXR \cite{johnson_mimic-cxr_2019}), or that have images and segmentation labels (e.g., SIIM Kaggle \cite{tolkachev_deep_2021}), there is a scarcity of large datasets containing all three elements. An additional limitation is that most of the samples in the CANDID-PTX dataset were labeled by a single physician. While the original CANDID-PTX study did analyze inter-observer variability, our model's pneumothorax segmentations likely reflect the tendencies of the primary annotator. 
 

In conclusion, we demonstrated the feasibility of using vision-language models to enhance medical image segmentation, and showed that descriptive language can guide medical image analysis algorithms.

\section*{Acknowledgments}

This research was supported by GE Healthcare
and by an NVIDIA Academic Hardware Grant.

\section{Declarations}

\subsection{Funding} Tyler Bradshaw received funding through a master research agreement from GE Healthcare.

\subsection{Competing Interests}
Tyler Bradshaw, Zachary Huemann, and Junjie Hu have a patent pending for Automatic Image Quantification From Physician-Generated Reports.
Zachary Huemann received Nvidia RTX6000 GPU as an Academic Hardware Grant in support of this project. 

\subsection{Ethics Approval}
Institutional Review Board approval was obtained. Informed consent was waived by IRB due to minimal risk to subjects

\bibliographystyle{unsrt}  
\bibliography{references}

\end{document}